# Going Headless? On the Boundaries of Vertical AI Firms

**Muhammad Zia Hydari**
University of Pittsburgh
hydari@alum.mit.edu
ORCID: 0000-0003-4522-326X

**Farooq Muzaffar**
Ensi.ai
fmuzaffar@alum.mit.edu
ORCID: 0009-0007-2310-8144

**Abstract**

Vertical AI firms, the companies that have built specialized AI products for accounting, law, healthcare, procurement, and similar domains, historically bundled workflow, domain logic, and accountability into a single application. General-purpose AI agents are now unbundling that package, prompting founders and investors to advocate that vertical AI firms should “go headless”: cede the workflow and interface to agents and expose domain expertise as callable services. This article argues that going headless is correct for some firms and destructive for others, and that firms in the second category often cede their value capture inadvertently through architectural choices that look like interface decisions. The strategic question is which parts of the bundle should remain inside the firm, and the answer turns on distinguishing the interface boundary, which can often move, from the accountability boundary, which often must not. Drawing on Coase’s theory of the firm, Eisenmann, Parker, and Van Alstyne’s platform envelopment framework, and Teece’s analysis of complementary assets and appropriability, the article shows that orchestrators operating through open protocols acquire envelopment power over vertical firms even as technical interoperability improves, and that durable value capture concentrates in cospecialized accountability assets: professional signoff, regulated workflows, evidence trails, and trusted systems of record. The article proposes a three-position taxonomy (component, integrated software platform, dual-track) determined not by sector but by task-accountability regime, and formalizes the construct of rule debt: the future governance, maintenance, and accountability burden that accrues to customer organizations when business rules and professional standards migrate from governed systems into prompts and agent instructions. Four operating principles follow: decompose by accountability rather than interface, invert the edges while retaining the core, position rule debt as the customer cost the integrated platform prevents, and avoid single-orchestrator dependence. An appendix demonstrates that Williamson’s hold-up framework and the Grossman, Hart, and Moore investment-incentive argument yield convergent prescriptions for the bilateral-contracting subset.

---

---

Vertical AI firms—the companies that have built specialized AI products for accounting, law, healthcare, procurement, software engineering, insurance, and similar domains—have historically bundled several activities into a single application. They connected to the customer's data and systems. They structured the workflow. They encoded domain-specific knowledge and decision logic. They generated outputs, managed approvals, and maintained the evidence trail that made those outputs defensible. The bundle was the product.

General-purpose AI agents are now beginning to unbundle it. When an agent can plan across files, call tools, manipulate documents, and coordinate multi-step tasks, several activities that used to require a vertical application can be performed inside the agent itself. This has prompted a strategic response that founders and investors have begun to advocate openly: vertical AI firms should go "headless." Derived from traditional software engineering, a headless architecture refers to a design paradigm where the frontend presentation layer (the user interface) is entirely decoupled from the backend logic and data storage, communicating exclusively via application programming interfaces (APIs). In this new context, going headless means that vertical AI firms should cede the workflow and interface to general-purpose agents and expose their domain expertise as callable services. The advice has real appeal beyond founder fashion. Orchestrators carry distribution to hundreds of millions of users an individual vertical firm cannot match alone. The prevailing architectural narrative treats workflow as a commoditizing layer that agents will absorb anyway. Capital markets reward "AI-native" architectures over vertical SaaS multiples. And offloading workflow assembly to a general agent saves engineering capacity that can be redirected to the domain capability itself. The question is therefore not whether the advice is appealing but for which firms it actually works.

That response is correct for some firms and destructive for others. In domains where customers buy accountable judgment rather than task completion, going headless can transfer the interface while dissipating the assets that made the vertical system valuable. It can also shift a new liability onto the customer organization: *rule debt*, the future governance burden that accrues when decision rules migrate from governed systems into prompts and agent instructions. The strategic question is not whether vertical AI firms will integrate with agents (almost every firm will) but which parts of the bundle should remain inside the firm and which should be unbundled. This is a boundary question, and the answer turns on distinguishing the interface boundary (which can often move) from the accountability boundary (which often must not). The danger is not that firms knowingly give away the assets that capture value. It is that the headless prescription makes accountability infrastructure look like ordinary workflow, so value capture migrates to whoever holds the interface (even when the underlying functionality stays at the vertical firm) before managers recognize that a boundary decision has been made.

## Why Agents Unbundle the Vertical AI Firm

Coase argued that firms exist because using markets is not costless [1]. Activities are internalized when searching, coordinating, monitoring, and contracting through the market cost more than managerial direction. Customers facing a need for AI-supported work in accounting, law, healthcare, or similar domains have always had three options: build the capability themselves, buy an integrated vertical AI product, or assemble multiple components from separate providers. The first option was rarely competitive, since most customers lacked the domain expertise, model access, data, and engineering capacity to match specialist providers, so the economically meaningful comparison was between buying the bundle and assembling components.[1] The Coasian comparison between those two options turned on a single variable: the coordination cost of stitching components together in the market versus the coordination cost already absorbed inside the vertical AI firm's bundle. Historically, market coordination cost was the higher of the two: integrating components, handling errors, reconciling schemas, and maintaining the connective glue typically exceeded the premium vertical AI firms charged to deliver the bundle pre-assembled.[2] So customers bought bundles, and vertical AI firms grew up around bundling. Agents change which side of the comparison is higher. By acting as the integrator the customer was previously unwilling to hire, the agent collapses the market coordination cost that previously made assembly uneconomic. The vertical firm's product boundary then moves as a derivative response: customers are willing to pay for the parts of the bundle that remain hard for the agent to assemble, and less willing to pay for the parts the agent can assemble cheaply. Different capabilities face different willingness-to-pay shifts, which is why the unbundling is uneven.

## Three Orchestrator Relationships

The market structure of orchestrator and vertical-firm relationships is worth naming, because it constrains which economic mechanisms actually apply. Three patterns coexist today: open protocols governed by public specifications, developer terms, or standardized access rules rather than negotiated, relationship-specific bilateral contracts (the long tail of Model Context Protocol servers, Gemini connectors, the open layer of OpenAI's Apps SDK); curated commercial partnerships with negotiated contracts but no revenue share; and explicit revenue-sharing arrangements.[3] The argument that follows applies most directly to the open-protocol pattern, which is the dominant case for the vertical AI firms the article is about. Frameworks built for bilateral contracting fit the curated and revenue-share patterns; [4] the open-protocol pattern requires platform-economic logic on platform openness and control instead [2].

---

[1]The build option was structurally noncompetitive because specialist providers had amortized the high fixed costs of building AI capabilities (domain expertise, model access, training data, evaluation infrastructure, and engineering capacity) across many customers, and the customer building alone would have had to bear those costs without amortization. This is the classical economics of specialization. The article focuses on the comparison between buying the bundle and assembling components because that is where agents change the economics, not because the build option is irrelevant in principle.

[2]These computational frictions correspond to the three categories of market coordination cost that Carl Dahlman formalized from Coase's work [Dahlman, C. J. 1979. The Problem of Externality. Journal of Law and Economics 22, 1, 141–162.], with two mapping exactly and one mapping approximately. Searching for the right component, which involves identifying which providers offer a needed capability, evaluating their quality, and comparing alternatives, is search cost in its original sense. Handling errors across a multi-component workflow, which involves verifying that each call returned a correct result, attributing failures when something goes wrong, and remediating, is enforcement cost in its original sense. Reconciling schemas, which involves agreeing on how data is structured, how fields are named, and how errors are signaled (for example, ensuring that one component's MM-DD-YYYY date output matches another's YYYY-MM-DD requirement before the two can communicate), corresponds approximately to bargaining cost. Modern API economics has standardized most commercial bargaining, leaving the technical interface contract as the residual form of inter-party agreement that must still be negotiated. The vocabulary is computational and the standardization is greater than in Coase's industrial setting, but the underlying friction structure is the one he identified.

[3]Curated partnerships include the Thomson Reuters CoCounsel integration with Anthropic and enterprise-tier integrations announced by Anthropic and OpenAI with established legal and finance vendors. Revenue-sharing examples include AWS, Azure, and Google Cloud marketplaces, which charge platform fees on partner-agent

## Why Open Protocols Do Not Remove Platform Power

Platform envelopment is the mechanism that operates in the open-protocol case. The platform actor in this analysis is the orchestrator, not the vertical AI firm: the orchestrator aggregates users, tools, data sources, and third-party capabilities into the surface where work is requested, routed, and completed. By coordinating the various components serving a customer's transactions, it assumes the aggregate platform position and the resulting envelopment power. Eisenmann, Parker, and Van Alstyne describe envelopment as the absorption of one platform's functionality into another by exploiting shared user bases and complementary capabilities, pushing the absorbed offering toward commoditization or irrelevance [3]. The dynamic maps onto agent ecosystems through three related mechanisms. The first is envelopment proper: the orchestrator builds native capabilities that substitute for the vertical tool, absorbing the functionality into its own offering. The second is platform gatekeeping: without building a substitute, the orchestrator changes which third-party tools its agent reaches for first, alters default tool selection, deprecates categories, or extends the protocol in ways that favor some complementors over others. The third operates over longer time horizons: knowledge distillation and product-learning advantages from interface control. Where data-use terms permit, the input-output traces of customer interactions can become training or fine-tuning data for the orchestrator's own model, supplying competitor training data through normal protocol operation.[5] Even where enterprise terms prohibit training on customer data, the orchestrator gains product-learning advantages from controlling the interface: it observes which tools are invoked, where workflows fail, which outputs users accept or revise, and which vertical functions attract repeated demand. Over many calls, these

---

transactions, and OpenAI's GPT earnings pilot. OpenAI's Instant Checkout, launched in 2025 with a small merchant fee on completed purchases, illustrates the instability of transaction-control models in agent ecosystems. OpenAI's current Apps SDK materials (https://perma.cc/9UTE-WSHR and https://perma.cc/872U-QRJ8) describe external checkout as the recommended and generally available path for most developers, while in-app checkout through the ChatGPT payment sheet remains limited to select marketplace partners or beta contexts. The Wolfram and OpenAI plugin of 2023 illustrates a fourth pattern: featured developer-program participation that the platform can sunset unilaterally, as OpenAI did when it deprecated plugins.

[4] For the bilateral-contracting patterns, similar conclusions follow from Williamson's hold-up framework and the Grossman, Hart, and Moore investment-incentive argument, with the additional protection that explicit contracts and residual control rights afford. Appendix A develops this argument, applying both frameworks to the curated-partnership and revenue-share subsets and demonstrating convergence with the main article's prescriptions through different mechanisms.

[5] In practice, enterprise customers increasingly negotiate data-use agreements that prohibit the orchestrator from using interaction data for model training, and the major orchestrator providers (including OpenAI, Anthropic, and Google) offer enterprise terms that exclude commercial or API inputs and outputs from training by default. These contractual protections limit the knowledge-distillation mechanism for large enterprise deployments. The mechanism remains operative, however, where such protections are absent: vertical AI firms or startups operating through non-business accounts, consumer-tier usage, shadow AI deployments in loosely managed small and mid-sized businesses, and any deployment under default terms that permit training. The practical boundary between operational observation of interaction patterns and learning from them is also difficult to verify externally even when contractual restrictions exist.

routing, evaluation, and telemetry advantages can help the orchestrator approximate, replace, or route around specialized components. All three mechanisms expose vertical AI firms to orchestrator decisions they cannot veto, and none requires a contract with the complementor; all three require only that the orchestrator owns the user-facing surface. Open protocols such as the Model Context Protocol reduce one form of friction by standardizing how agents connect to external tools and data [4], but they do not constrain the orchestrator's residual power to absorb, re-rank, substitute, or distill. The interface is portable; the economic position may not be. The more a vertical AI firm optimizes for one orchestrator's ranking, evaluation habits, permission model, or protocol extensions, the more exposed it becomes to that orchestrator's envelopment choices.

## Complementary Assets and Accountability

Teece completes the argument [5]. Innovation profits often accrue not to the inventor but to whoever controls the complementary assets needed to commercialize the invention. Complementary assets are the resources that turn an invention into a deliverable product: manufacturing capacity, distribution channels, service networks, brand, regulatory approvals, customer relationships, and the operational infrastructure that supports continued use. An invention without its complementary assets is a laboratory result, not a business. EMI's CT scanner is the classic example: EMI developed the breakthrough, but firms such as GE and Siemens were better positioned to scale the business through manufacturing, hospital sales, and service networks. The breakthrough belonged to EMI; the value belonged to whoever owned the complementary assets.

Teece distinguishes three kinds of complementary assets, and the distinction is what determines who captures the value. Generic complementary assets are widely available and confer no particular advantage; anyone can rent them. Specialized complementary assets exhibit one-way dependence: the asset depends on the innovation but the innovation could in principle work with substitute assets. Cospecialized complementary assets exhibit two-way dependence: the asset depends on the innovation and the innovation depends on the asset, with no easy substitute on either side. The cospecialized category is where strategic outcomes diverge, because cospecialization creates mutual dependence that binds value capture to whoever owns the asset.

In vertical AI, generic assets such as compute confer little advantage because anyone can rent compute. Specialized assets such as proprietary domain data tilt advantage toward whoever owns them, but the AI capability could in principle work with alternative data sources. The cospecialized assets are the ones that bind directly to the AI-enabled offering and bind the AI-enabled offering to them: professional signoff (a licensed CPA or attorney whose name attaches to the conclusion the AI helped produce), regulated workflows (audit procedures, clinical protocols, or compliance processes that define what the AI's output means), evidence trails (the documentation chain that makes an AI-supported conclusion defensible to a regulator or court), and trusted systems of record (the authoritative repositories whose contents AI outputs must be reconciled against and written back into). The AI capability makes these assets more scalable and efficient, while these assets give the AI capability its legitimacy, defensibility, and operational integration.

Control over cospecialized assets materially improves a firm's ability to appropriate value, and that is where boundary choices over cospecialized assets become consequential.

Teece's framework adds a second variable that interacts with complementary-asset ownership: appropriability. Appropriability refers to how easily an innovator can prevent others from imitating the innovation itself, separate from any complementary asset. Strong appropriability comes from patents, secrecy, tacit knowledge, or lead-time advantages that competitors cannot quickly close; weak appropriability means the innovation can be freely replicated. When appropriability is strong, the innovator can capture value even without owning the full complementary-asset stack, because nobody else can produce the innovation. When appropriability is weak, complementary-asset ownership becomes the dominant source of durable value capture, because the innovation itself confers no protection. The argument for owning cospecialized assets is therefore sharpest where appropriability around the AI capability is weak, where a frontier model can closely approximate the capability with ordinary prompting. Where appropriability is strong, through proprietary fine-tuning data that competitors cannot replicate, accumulated evaluation sets tightly coupled to the model, or genuinely tacit domain knowledge embedded in deployment, the firm has additional optionality and may capture value without owning the full accountability stack. Many vertical AI capabilities, especially those based on generic reasoning over documents, structured extraction, or first-pass summarization, sit closer to the weak-appropriability end, which is why cospecialized assets matter.

The investment behavior of firms is shaped by the same logic. Firms invest in assets whose returns they expect to capture and avoid investing in assets whose returns will accrue to someone else. In agent ecosystems, this creates a clean asymmetry between orchestrators and vertical firms, driven by how substitutable their investments are across domains. An orchestrator's investments in planning quality, tool routing, and general reasoning capability are highly substitutable across domains: the same planning engine serves accounting, law, healthcare, and a hundred other applications, so the orchestrator amortizes those investments across an enormous customer base and has strong incentive to make them. A vertical firm's investments in evidence systems, review processes, and escalation paths are not substitutable across domains: audit evidence infrastructure does not transfer to clinical decision support, and a clinical workflow is useless to an auditor. The vertical firm accumulates these domain-specific assets slowly through years of client and regulatory engagement, and only it has reason to invest in them. The boundary between orchestrator and vertical firm should therefore track this substitutability gradient. Where the customer's combined output depends most on general planning capability, the orchestrator is the rational investor and owner. Where the customer's combined output depends most on domain-specific accountability infrastructure, the vertical firm is the rational investor and owner or controller; under arrangements that leave those assets controlled elsewhere, no party has the incentive to invest in them at the level the domain requires, and the accountability infrastructure deteriorates. Most accountability-bound domains fall into the second category, which is why the vertical firm's boundary should retain ownership of cospecialized assets even when the interface boundary moves.

These frameworks address different parts of the same boundary problem. Coase explains when the bundle unravels under falling assembly costs. Platform envelopment explains why ceding the interface to an orchestrator creates strategic dependence even when technical interoperability improves. Teece explains why control over evidence, signoff, and governed workflows shapes value capture. And rule debt, developed below, names the operational cost of letting decision logic migrate out of governed systems. Together they identify three boundaries a vertical AI firm must distinguish: the interface boundary (where the product meets the orchestrator), the value-capture boundary (where appropriability concentrates), and the accountability boundary (where responsibility, evidence, and governance must reside). The first can often move; the third typically should not.

## Three Strategic Positions

This yields three strategic positions: component, integrated software platform,[6] and dual-track. The distinction is not by sector but by task-accountability regime. A healthcare firm, a legal technology firm, or a procurement platform may contain all three at once. The error is to classify the company by its market label rather than each capability by its economics. A capability is a component when verification is cheap and responsibility is transferable. It is platform work when context, evidence, and authority must be preserved. It is dual-track when task execution can be modularized but the final commitment cannot. These positions are contingent, not permanent. Capabilities can migrate between them as outputs standardize, verification becomes statistical, regulators adapt, and customers accept new allocations of responsibility — the dynamic the article returns to below. A capability that is platform work today may be dual-track in three years and a component in seven. The static taxonomy is a diagnostic snapshot, not a prediction.

First, some vertical AI businesses should become components. Their outputs are modular, low-stakes, and cheap to verify. Examples include semi-structured data extraction, document classification, standards retrieval, and first-pass summarization. Document-intelligence vendors such as Hyperscience and Rossum (the latter acquired by Coupa in May 2026) built their original businesses around bundled extraction and classification platforms for finance and operations workflows, but increasingly illustrate how core document-intelligence capabilities can be exposed as callable components inside larger enterprise systems. The component-tier business does not require foundation models to remain weak at parsing; it requires only that the accuracy gap between general and specialized infrastructure remain meaningful in high-stakes document workflows. The customer often does not want another standalone application. The customer wants the capability available inside an agent, an enterprise workflow, or an internal system.

For component businesses, headless is distribution. The orchestrator is the channel through which customers find and invoke the firm's capability, replacing the standalone product that previously served the same function. Their priorities are reliability, latency, accuracy, price, schemas, permissions, and broad integration. The danger is not becoming a component. The danger is becoming a commodity component. Where the accuracy gap closes, and a frontier model can closely approximate the capability with ordinary prompting, the component must create defensibility through proprietary feedback loops, benchmarking, data access, trust, or scale.

[6]We use "integrated software platform" in the architectural sense (an end-to-end software product bundling workflow, domain logic, and accountability within a single firm boundary), not in the multi-sided platform sense used in the envelopment discussion above. In the multi-sided sense, the orchestrator is the platform actor, as the envelopment section establishes; the vertical firm operating as an integrated software platform is a complementor whose product happens to bundle several activities.

Second, some vertical AI firms should remain integrated software platforms because they sell accountable judgment: outputs whose value depends not only on correctness but on traceability, authority, reviewability, and the ability to assign responsibility after the fact. Here errors are costly, verification is difficult, context accumulates over time, and responsibility cannot be cleanly transferred to a general agent. Accounting illustrates the point. A general agent may draft a reconciliation or extract invoice data. But an audit conclusion depends on evidence, materiality, prior judgments, review, documentation, and signoff. Audit documentation standards treat the evidence trail not as administrative residue, but as the basis for review and support for significant conclusions. Audit evidence standards also emphasize that reliability depends on the nature and source of evidence and the circumstances under which it is obtained [6, 7].

The same logic appears beyond licensed professions. In industrial maintenance, a system may predict equipment failure, but the decision to defer maintenance on a turbine, refinery component, or aircraft subsystem carries safety, operational, and regulatory exposures that must attach to a named accountable party.[7] These are accountability architectures, not merely workflows. A vertical AI firm serving such domains must therefore remain an integrated software platform, or at least retain enforceable control over the accountability layer, because the relevant cospecialized assets (the maintenance decision record, the named-party signoff infrastructure, the evidence chain that supports defensible operating decisions) cannot be reassembled by an agent at runtime from external components.

For integrated software platforms, integration with agents is necessary but disappearance into agents is dangerous. The platform should let agents initiate work, call bounded functions, and retrieve status. It should not surrender the system of record, system of evidence, or system of responsibility.

Disappearance into agents is rarely a deliberate choice; it is the foreseeable consequence of an architectural decision made for other reasons. A firm that exposes its capability as a callable tool while ceding the workflow may believe it has retained the domain capability while shedding only the interface. In practice, the orchestrator now controls the user relationship, the evaluation surface, the audit trail of which tools were invoked, and the surface where signoff occurs. The cospecialized assets that made the capability defensible accumulate at whoever holds the interface, not at whoever holds the underlying capability. A firm following the headless prescription in an accountability-bound domain may discover that it has not just moved its interface to the orchestrator but given away the value-capture surface, and the discovery typically comes only after the migration is complete and the assets have already begun to dissipate.

[7]As of May 2026, vertical AI firms operating in industrial maintenance include C3 AI (whose C3 AI Reliability application is deployed by Shell, Dow, Georgia-Pacific, and the US Air Force for asset monitoring at scale), Augury (machine-health monitoring partnered with Baker Hughes through the Bently Nevada brand), and Uptake Technologies (predictive maintenance for commercial fleets, whose planned acquisition by Bosch was announced in March 2026). The named accountable party in each case is the operator of the asset; the AI firm's product is the evidence and recommendation infrastructure that supports the operator's decision.

The metaphor itself is revealing: a headless firm is also an unaccountable one. The word founders use to celebrate distribution also names its failure mode in accountability-bound domains.

Third, many firms are dual-track. They can expose components at the task boundary while preserving platforms at the accountability boundary. Legal research, citation checking, and document comparison can be components; a filed brief or legal opinion remains governed work. Healthcare summarization and coding can be components; diagnosis and care planning require clinical governance. Software generation can be modular; production release, security approval, and incident responsibility remain platform-like.

Dual-track firms have two businesses, not one. The component business optimizes for adoption, usage, interoperability, latency, and gross margin. The platform business optimizes for retention, governance, risk reduction, evidence, and trust. Treating them as one business prices accountability like traffic and evaluates governed work like an API call.

## Rule Debt as the Customer-Side Cost of Headlessness

The hidden cost of headless automation, for the customer organization adopting it, is rule debt.[8] Rule debt is the future governance, maintenance, and accountability burden created when business rules, professional standards, or operating policies are encoded in prompts, saved tasks, ad hoc scripts, and agent instructions rather than in governed systems. The term intentionally echoes technical debt, the metaphor Ward Cunningham introduced to describe how expedient software choices create future rework [8]. But rule debt is more specific: it is debt in the organization's decision logic, not in its code. It is also distinct from related concepts: traditional technical debt lives in code, model risk addresses model behavior in isolation, and shadow IT describes ungoverned tooling. Rule debt is the migration of organizational decision rules from governed systems into informal agent instructions that are hard to inventory, test, version, reconcile, and assign to accountable owners. A prompt that tells an agent how to classify a transaction, escalate a claim, summarize a medical message, assess a supplier, or approve an exception is not just an instruction. It is an informal policy document. When the policy changes, the prompt must be found, updated, tested, versioned, and reconciled with other prompts. At small scale, this feels flexible. At organizational scale, it becomes an ungoverned control environment.

Enterprise experience confirms the pattern: organizations scaling agentic AI encounter recurring governance frictions around agent identity, data provenance, probabilistic control, and accountability that parallel workforce management rather than software configuration [12].

The point is not that prompts or agent instructions should be avoided. They are often the fastest way to adapt work to local context. The point is that, once they encode business rules or professional standards, they need governance: ownership, versioning, testing, monitoring, and reconciliation with the organization's formal policies.

---

[8]The phrase has been used loosely in industry writing to describe maintenance burdens of legacy rule-based systems. This article gives it a sharper, agent-era meaning.

## When Accountability Migrates

The accountability moat is not absolute. Some activities once treated as integrated professional judgment have been standardized and unbundled. Technology-assisted review made large-scale legal document review more tool-mediated. Robo-advisers brought algorithmic portfolio management into a regulated advisory setting, while remaining subject to fiduciary obligations as registered investment advisers. Accountability can migrate when outputs become standardized, verification becomes statistical, regulators accept new control frameworks, and customers accept a different allocation of responsibility [9, 10].

That is the central migration problem for incumbents. Many vertical AI firms were built around owning workflow. Their architectures and sales processes assumed that work would happen inside their product. Moving from platform to component is therefore not a simple API launch. It can cannibalize the interface, confuse buyers, and expose that the prior moat was orchestration rather than judgment. The appropriability logic predicts the resulting resistance: firms underinvest in capability transitions that devalue the assets their prior ownership rewarded, even when the transition is rational for the broader economy.

## Operating Principles

Four operating principles follow.

First, decompose by accountability, not by user interface. Ask which tasks can be verified cheaply, which require accumulated context, and which create commitments that someone must defend. The modular parts should be callable. The accountable parts should remain governed.

Second, invert the edges, not the core. Platform research shows that APIs and developer ecosystems can move value creation outside the firm, but openness also changes control and risk [11]. Vertical AI firms should expose tools, data access, and task capabilities to external agents while retaining the assets that make final outputs reliable: permissions, evidence, evaluation, review, lineage, and signoff.

Third, position rule debt as the customer cost the integrated platform prevents. Customer organizations adopting agentic automation accumulate rule debt when business rules, professional standards, and operating policies migrate into prompts and agent instructions rather than into governed systems. Vertical AI firms in accountability-bound domains can offer the integrated platform as the structural prevention of this debt. Firms operating as components in such domains should at minimum provide the governance tooling (instruction registries, prompt versioning, adversarial testing, behavior logging, exception ownership) that customer organizations need to manage rule debt themselves.

Fourth, avoid single-orchestrator dependence. A headless strategy bound to one agent recreates the envelopment exposure it was meant to avoid. Vertical AI firms operating in the open-protocol pattern should support multiple orchestrators and preserve direct customer relationships where accountability matters.

The vertical AI firms that survive will not defend every screen, nor will they dissolve themselves into someone else's agent. They will make their expertise easy to call and their accountability hard to copy. The losers will celebrate going headless while quietly giving up the assets that made them vertical, and the surrender will show up not in their announcements but in their underinvestment in the things they no longer own or control.

## References

[1] Coase, R. H. 1937. The Nature of the Firm. Economica 4, 16, 386–405. https://www.jstor.org/stable/2626876

[2] Boudreau, K. 2010. Open Platform Strategies and Innovation: Granting Access vs. Devolving Control. Management Science 56, 10, 1849–1872. https://doi.org/10.1287/mnsc.1100.1215

[3] Eisenmann, T., Parker, G., and Van Alstyne, M. 2011. Platform Envelopment. Strategic Management Journal 32, 12, 1270–1285. https://doi.org/10.1002/smj.935

[4] Model Context Protocol. 2025. Specification. https://modelcontextprotocol.io/specification/2025-11-25

[5] Teece, D. J. 1986. Profiting from Technological Innovation: Implications for Integration, Collaboration, Licensing and Public Policy. Research Policy 15, 6, 285–305. https://doi.org/10.1016/0048-7333(86)90027-2

[6] Public Company Accounting Oversight Board. AS 1215: Audit Documentation. https://pcaobus.org/oversight/standards/auditing-standards/details/AS1215

[7] Public Company Accounting Oversight Board. AS 1105: Audit Evidence. https://pcaobus.org/oversight/standards/auditing-standards/details/AS1105

[8] Cunningham, W. 1992. The WyCash Portfolio Management System. OOPSLA ’92 Experience Report. https://dl.acm.org/doi/10.1145/157709.157715

[9] American Bar Association. 2023. Predictive Coding Technologies for the Modern Lawyer. Litigation News. https://www.americanbar.org/groups/litigation/resources/litigation-news/2023/summer/predictive-coding-technologies-modern-lawyer/

[10] U.S. Securities and Exchange Commission, Division of Investment Management. 2017. Robo-Advisers. IM Guidance Update No. 2017-02. https://www.sec.gov/investment/im-guidance-2017-02.pdf

[11] Parker, G., Van Alstyne, M., and Jiang, X. 2017. Platform Ecosystems: How Developers Invert the Firm. MIS Quarterly 41, 1, 255–266. https://aisel.aisnet.org/misq/vol41/iss1/15/

[12] Telang, R., Hydari, M. Z., and Iqbal, R. 2026. To Scale AI Agents Successfully, Think of Them Like Team Members. Harvard Business Review (March 2026). https://hbr.org/2026/03/to-scale-ai-agents-successfully-think-of-them-like-team-members

## Appendix A. Bilateral Contracting in Vertical AI: What Changes When There Is a Partnership?

The main article focuses on the open-protocol pattern, in which a vertical AI firm connects to an orchestrator through public protocols, developer terms, or standardized access rules rather than through a negotiated, relationship-specific contract. In that setting, platform envelopment and complementary-assets logic are the most direct frameworks.

Some vertical AI relationships are different. A legal AI firm may enter a curated partnership with a general-purpose agent provider. An accounting or procurement AI firm may join a marketplace in which the platform takes a fee or shares revenue. In these cases, the orchestrator and the vertical firm do have a bilateral contract. The contract can define integration obligations, support commitments, commercial terms, data-use restrictions, co-marketing rights, and termination procedures.

That contract matters. It gives the vertical firm more protection than an open-protocol participant has. But it does not eliminate the boundary problem. The orchestrator still usually owns the user-facing surface, the routing logic, the ranking rules, the evaluation criteria, and the terms under which third-party capabilities are discovered. The vertical firm still has to decide which activities to expose to the orchestrator and which accountability assets to keep under its own control.

Two classic economic ideas clarify this bilateral-contracting case.

### *Williamson: Do Not Customize Yourself Into Dependence*

Williamson's hold-up framework begins with a simple observation: some investments are much more valuable inside a particular relationship than outside it. Once those investments are made, the counterparty may have leverage, because the investing firm cannot easily walk away without losing much of the value it has created.

In vertical AI partnerships, this is easy to see. A vertical firm may customize its API to one orchestrator's tool-calling format, tune its schemas to that orchestrator's planner, adapt its permission model to the orchestrator's identity system, assign a partner-management team, join a co-marketing program, or optimize its product for the orchestrator's ranking and discovery surface. These investments may be valuable inside the relationship but much less valuable elsewhere.

After those investments are sunk, the orchestrator may not need to renegotiate the headline contract to change the economics of the relationship. It can alter default tool selection, change ranking criteria, adjust eligibility rules, modify protocol requirements, bundle a competing first-party capability, or route more traffic to another partner. In a revenue-sharing arrangement, the formal percentage split may remain unchanged, while the effective value of the relationship changes through traffic, attribution, refund rules, bundling, or placement.

The Williamson lesson is therefore practical: vertical AI firms should be cautious about investments that are valuable only inside one orchestrator relationship. They should negotiate protections where possible, including minimum terms, API-deprecation notice, change-management obligations, attribution rules, data-use restrictions, audit rights, service-level commitments, and transition rights at termination. But the more important protection is strategic rather than legal: invest most heavily in assets that remain valuable across orchestrators and across direct customer relationships.

For vertical AI firms, those assets are usually the accountability assets: evidence trails, evaluation infrastructure, professional review processes, escalation paths, regulated workflows, proprietary domain data, customer trust, and systems-of-record integrations. These assets may support an orchestrator partnership, but they do not depend on any single orchestrator for their value.

### *Grossman-Hart-Moore: Control Determines Investment*

The Grossman-Hart-Moore framework starts from another simple observation: contracts cannot specify everything that may happen in the future. A partnership contract may say how an integration works today, but it cannot fully specify what happens if the orchestrator changes its ranking logic, launches a competing native feature, alters its evaluation method, changes its user interface, or extends the protocol in a way that favors some partners over others.

When the contract is silent, control over the relevant asset matters. The orchestrator usually controls the interface, the marketplace, the routing system, the user relationship, and the rules by which third-party tools are surfaced. The vertical firm controls, or should try to control, the domain-specific assets that make its outputs trustworthy: the evidence chain, the review workflow, the audit log, the professional signoff process, the governed decision rules, and the customer's accountable workflow.

This matters because firms invest more in assets whose returns they expect to capture. If the vertical firm's returns depend mainly on the orchestrator's future routing and ranking choices, the vertical firm has weaker incentives to invest deeply in that relationship. If, instead, the vertical firm owns or controls the accountability assets that remain valuable regardless of which orchestrator is used, it has stronger incentives to keep improving them.

This is especially important in accountability-bound domains. A general-purpose agent may initiate an audit task, request a contract comparison, summarize a clinical note, or call a procurement-risk API. But the final accountable output still depends on assets that the orchestrator does not usually own: documentation, review, escalation, lineage, permissions, professional judgment, and a defensible record of why the decision was made. Those assets should remain under the control of the vertical firm, the customer organization, or the regulated professional context in which the decision becomes binding.

### *What Bilateral Contracts Add, and What They Do Not*

Bilateral contracting improves the vertical firm's position relative to the open-protocol case. A curated partner can negotiate enforceable terms. A revenue-sharing partner can secure commercial participation in the orchestrator's success. Both can obtain more notice, more predictability, and more formal remedies than an ordinary open-protocol participant.

But bilateral contracting does not change the central prescription of the article. It changes the degree of protection, not the underlying boundary logic.

A vertical AI firm should expose the functions that benefit from agent distribution: retrieval, drafting, extraction, classification, comparison, routing, and other callable tasks. It should avoid tailoring itself so narrowly to one orchestrator that its capabilities lose value elsewhere. Most importantly, it should retain control over the accountability assets that make its outputs defensible: evidence, review, evaluation, lineage, permissions, governed workflows, professional signoff, and customer trust.

The open-protocol case, the curated-partnership case, and the revenue-sharing case therefore differ in legal structure but converge on the same strategic lesson. The interface can move toward the orchestrator. The accountability boundary should move only when accountability itself has migrated.

In Williamson's terms, the vertical firm should avoid creating relationship-specific dependence that allows the orchestrator to capture the value of sunk investments. In Grossman-Hart-Moore terms, the vertical firm should own or control the assets that determine its incentive to invest. In the main article's terms, the firm should make its expertise easy to call and its accountability hard to copy.

## Appendix References

[13] Williamson, O. E. 1985. The Economic Institutions of Capitalism: Firms, Markets, Relational Contracting. Free Press.

[14] Grossman, S. J. and Hart, O. D. 1986. The Costs and Benefits of Ownership: A Theory of Vertical and Lateral Integration. Journal of Political Economy 94, 4, 691–719.

[15] Hart, O. and Moore, J. 1990. Property Rights and the Nature of the Firm. Journal of Political Economy 98, 6, 1119–1158.